\documentclass{article}

\usepackage{microtype}
\usepackage{graphicx}
\usepackage{subfigure}
\usepackage{booktabs} 

\usepackage{hyperref}

\usepackage{cite}
\usepackage{amsmath,amssymb,amsfonts}
\usepackage{graphicx}
\usepackage{textcomp}
\usepackage{xcolor}
\usepackage{lipsum}
\usepackage{hyperref}
\usepackage{float}

\usepackage[accepted]{icml2025}

\usepackage{amsmath}
\usepackage{amssymb}
\usepackage{mathtools}
\usepackage{amsthm}

\usepackage[capitalize,noabbrev]{cleveref}

\theoremstyle{plain}

\theoremstyle{definition}

\theoremstyle{remark}

\usepackage[textsize=tiny]{todonotes}

\icmltitlerunning{Token-Efficient RL for LLM Reasoning}

\begin{document}

\twocolumn[
\icmltitle{Token-Efficient RL for LLM Reasoning}



\icmlsetsymbol{equal}{*}

\begin{icmlauthorlist}
\icmlauthor{Alan Lee}{equal,yyy}
\icmlauthor{Harry Tong}{equal,yyy}
\end{icmlauthorlist}

\icmlaffiliation{yyy}{Department of Computer Science and Engineering, University of Michigan, Ann Arbor, USA}

\icmlcorrespondingauthor{Alan Lee}{alanleee@umich.edu}
\icmlcorrespondingauthor{Harry Tong}{harryt@umich.edu}

\icmlkeywords{Machine Learning, ICML}

\vskip 0.3in
]



\printAffiliationsAndNotice{\icmlEqualContribution} 

\begin{abstract}
We propose reinforcement learning (RL) strategies tailored for reasoning in large language models (LLMs) under strict memory and compute limits, with a particular focus on compatibility with LoRA fine-tuning. Building on early policy gradient methods with baseline subtraction, we design critic-free methods that operate on a small, informative subset of output tokens to reduce memory usage and stabilize training. We introduce S-GRPO, a stochastic variant of Group Relative Policy Optimization, and T-SPMO, a token-level prefix matching approach for fine-grained credit assignment. Applied to Qwen2-1.5B, our methods raise accuracy on the SVAMP benchmark from 46\% to over 70\% and show strong performance on multi-digit multiplication. Surprisingly, full-token GRPO under LoRA fails to improve over the base model, suggesting that selective token-level optimization may act as an implicit regularizer in low-parameter training regimes.
\end{abstract}

\section{Introduction}
Large Language Models (LLMs) have demonstrated substantial improvements in mathematical and structured problem-solving through reinforcement learning (RL) fine-tuning on specific tasks \cite{o1}. However, standard RL approaches such as Proximal Policy Optimization (PPO) \cite{ppo} and Group Relative Policy Optimization (GRPO) \cite{grpo} can be impractical for researchers operating under limited computational budgets, for the following reasons: (1) both methods compute loss over the entire output trajectory, (2) they are typically designed for full model fine-tuning, and (3) PPO additionally requires a critic network, further increasing memory demands.

In this work, we design and evaluate RL algorithms that are explicitly tailored to low compute constraints: Partial parameter updates, limited GPU capacity, and the need to mitigate overfitting in low-capacity adaptation settings. First, we introduce Stochastic-GRPO (S-GRPO), a lightweight variant of GRPO that samples tokens from output trajectories to contribute to the loss. Second, we propose Token-Specific Prefix Matching Optimization (T-SPMO), which assigns credit at a token granularity. Our methods extend early baseline-based policy gradients \citep[e.g.,][]{buy4,reinforce_williams} to a fine-grained token-level setting under LoRA constraints.

To evaluate these methods, we fine-tune the open-weight Qwen2-1.5B model \cite{qwen2} using LoRA \cite{lora} on two reasoning benchmarks: the SVAMP dataset \cite{svamp} and a multi-digit multiplication task. Despite updating the policy using only 30–50\% of generated tokens in S-GRPO and under 5\% in T-SPMO, both methods substantially improve test-set accuracy across both benchmarks. We also establish a full-token GRPO baseline under LoRA, which surprisingly did not improve performance from the base model. These results suggest that carefully designed, memory-efficient RL algorithms can significantly improve LLM reasoning under compute constraints—even outperforming exhaustive full-token optimization strategies.

\section{Related Work}

\subsection{RL for Language Models} Reinforcement learning has been widely used to align language model outputs with task-specific or user-defined objectives. Many RL algorithms used to fine-tune LLMs, such as PPO~\cite{ppo} and ReST-MCTS~\cite{rest-mcts}, require training secondary models, such as critic or process reward models, increasing compute demands. Group Relative Policy Optimization (GRPO)~\cite{grpo} removes the critic by using group-level reward normalization, yet still relies on full-model fine-tuning and full-sequence loss computation, limiting its applicability under compute constraints. 

\subsection{Baselines and Grouped Advantage Estimation} Classic policy gradient methods reduce variance by subtracting a baseline from the return~\cite{reinforce_williams}. Later work~\cite{buy4} demonstrated that using average rewards over sampled completions—rather than a learned value function—can yield strong performance. We extend this idea to token-level credit assignment for LLMs.

\subsection{Memory-Efficient Fine-Tuning} Techniques like Low-Rank Adaptation (LoRA)~\cite{lora} enable parameter-efficient adaptation of large models on modest hardware by training low-rank adapters in frozen networks. We focus on RL methods that are compatible with LoRA.

\section{Methods}
\subsection{Problem Setup}

We focus on enhancing the reasoning capabilities of large language models (LLMs) using RL under constrained computational resources. 

The task is framed as \textit{conditional generation with reward feedback}, where the model generates a sequence of tokens $Y = (y_1, y_2, \dots, y_T)$ in response to a prompt $x$, and receives a scalar reward signal $r(Y) \in \mathbb{R}$ based on the correctness or quality of the generated sequence.

Let $\pi_\theta(y_t \mid x, y_{<t})$ denote the autoregressive policy defined by the language model with parameters $\theta$. The objective is to optimize $\theta$ to maximize the expected reward over the distribution of completions:
\begin{equation}
\mathcal{J}(\theta) = \mathbb{E}_{Y \sim \pi_\theta} \left[ r(Y) \right].
\end{equation}

Traditional policy gradient methods often rely on estimating the advantage function using a separate value network, which can be computationally intensive and memory-demanding. To address these challenges, the recently popularized critic-free RL algorithm \textit{Group Relative Policy Optimization (GRPO)} estimates advantages by comparing the rewards of multiple responses generated for the same prompt \cite{grpo}.

For each input prompt, GRPO generates a group $\mathcal{G}$ completions from the current policy $\pi_\theta$. Each completion $y^{(i)}$ in group $i$ is tokenized as $y^{(i)} = (y^{(i)}_1, y^{(i)}_2, \dots, y^{(i)}_{T_i})$ and assigned a scalar reward $R_i$.

The advantage of each token is normalized using the group-level statistics:
\begin{equation}
\hat{A}_{i,t} = \frac{R_i - \mu_R}{\sigma_R}
\end{equation}
Where $\mu_R$ and $\sigma_R$ are the mean and standard deviation of rewards across the group.

The policy is updated to increase the likelihood of responses with positive advantages and decrease it for those with negative advantages:

\begin{equation}
\begin{aligned}
\mathcal{J}(\theta)
&= \frac{1}{|\mathcal{G}|}\sum_{i=1}^{|\mathcal{G}|}
   \frac{1}{|y^{(i)}|}\sum_{t=1}^{|y^{(i)}|}
   \Biggl\{
   \min\Biggl[
        \frac{\pi_\theta\!\bigl(y_t^{(i)}\!\mid\! x,y_{<t}^{(i)}\bigr)}
             {\pi_{\theta_{\text{old}}}\!\bigl(y_t^{(i)}\!\mid\! x,y_{<t}^{(i)}\bigr)}
        \,\hat{A}_{i,t}, \\[2pt]
&\qquad\text{clip}\!\Bigl(
          \frac{\pi_\theta\!\bigl(y_t^{(i)}\!\mid\! x,y_{<t}^{(i)}\bigr)}
               {\pi_{\theta_{\text{old}}}\!\bigl(y_t^{(i)}\!\mid\! x,y_{<t}^{(i)}\bigr)},
          1-\epsilon,\,1+\epsilon
        \Bigr)\hat{A}_{i,t}
      \Biggr] \\[4pt]
&\quad -\,\beta\,\mathbb{D}_{\text{KL}}\!\bigl[\pi_\theta \,\|\, \pi_{\text{ref}}\bigr]
   \Biggr\}
\end{aligned}
\label{eq:grpo}
\end{equation}

 Where $\pi_{\text{ref}}$ and $\pi_{\text{old}}$ are previous versions of $\pi_{\theta}$, which are updated to $\pi_{\theta}$ after varying iterations. $\epsilon$ is a hyperparameter for PPO-style objective clipping, and $\beta$ is a hyperparameter coefficient for the KL-divergence regularization. GRPO estimates the KL-Divergence using the unbiased estimator:

\begin{align}
\mathrm{D}_{\mathrm{KL}}\left[ \pi_\theta \, \| \, \pi_{\text{ref}} \right]
&= \frac{\pi_{\text{ref}}(y^{(i)}_t \mid x, y^{(i)}_{<t})}
        {\pi_\theta(y^{(i)}_t \mid x, y^{(i)}_{<t})} \notag \\
&\quad - \log\frac{\pi_{\text{ref}}(y^{(i)}_t \mid x, y^{(i)}_{<t})}
              {\pi_\theta(y^{(i)}_t \mid x, y^{(i)}_{<t})} - 1
\end{align}

To further accommodate limited computational resources, we employ \textit{Low-Rank Adaptation (LoRA)} \cite{lora} for parameter-efficient fine-tuning. LoRA allows us to fine-tune only a small subset of the model's parameters, significantly reducing memory requirements.

We evaluate our methods on two representative reasoning tasks:
\begin{itemize}
    \item \textbf{SVAMP}, a benchmark of verbal arithmetic problems requiring symbolic and numerical reasoning \cite{svamp}.
    \item \textbf{Multi-digit multiplication}, where models must learn step-wise calculation strategies.
\end{itemize}

\subsection{S-GRPO: Stochastic Group Relative Policy Optimization}

Stochastic Group Relative Policy Optimization (S-GRPO) extends GRPO to a low-memory setting by reducing the tokens that contribute to the gradient from the full response trajectory.

We also set the number of updates per problem (denoted as $\mu$ in the original GRPO paper \cite{grpo}) to 1, aligning with the default setting in lightweight RL libraries such as TRL. This choice both simplifies optimization by eliminating the need for PPO-style clipped objectives and reduces computational overhead, consistent with our focus on efficient fine-tuning. Also in alignment of the default setting in TRL, we do not update  $\pi_{\text{ref}}$ to $\pi_{\theta}$.

The policy objective is thus:
\begin{align}
\mathcal{J}_{\text{SGRPO}}(\theta)
&= \frac{1}{|\mathcal{G}|}
\sum_{i=1}^{\mathcal{G}}
\frac{1}{|\mathcal{T}_i|}
\sum_{t\in\mathcal{T}_i}(
\frac{\pi_\theta(y^{(i)}_t\mid x,y^{(i)}_{<t})}
     {\pi_\theta(y^{(i)}_t\mid x,y^{(i)}_{<t})\big|_{\text{no grad}}}
\hat{A}_{i,t} \notag \\
&\quad -\,\beta\,D_{\text{KL}}\!\left[\pi_\theta\;\|\;\pi_{\text{ref}}\right])
\label{eq:sgrpo}
\end{align}

Here, $\mathcal{T}_i$ denotes the set of tokens in completion $i$ that contribute to the loss. Tokens are included in $\mathcal{T}_i$ according to:
\begin{equation}
\begin{aligned}
t \in \mathcal{T}_i \iff
\begin{cases}
1 & \text{if } t < \alpha \\
\text{Bernoulli}(P) & \text{if } t \geq \alpha \text{ and } |\mathcal{T}_i| < k \\
0 & \text{otherwise}
\end{cases}
\end{aligned}
\label{eq:token_selection}
\end{equation}
where $\alpha$ is a cutoff index ensuring early tokens are always included, $k$ is the maximum number of tokens contributing to the loss, and $P$ is the probability of including later tokens stochastically.

This hybrid rule ensures early tokens—which often contain key semantic steps—are reliably updated, while sampling later tokens only when there is enough space. This strikes a balance between learning stability and memory efficiency.

Note that in GRPO, all tokens  $t$ are guaranteed to be in $\mathcal{T}_i$ \cite{grpo}.

\subsection{T-SPMO: Token-Specific Prefix Matching Optimization}

Token-Specific Prefix Matching Optimization (T-SPMO) is a token-level RL algorithm that enables fine-grained credit assignment without trajectory-wide statistics. Similarly to GRPO, we sample $|\mathcal{G}|$ completions per prompt. Inspired by grouped baselines in REINFORCE \cite{buy4, reinforce_williams}, we apply advantage estimation to token-level prefix transitions. We construct a token prefix trie from $\mathcal{G}$ and find all unique $(p, v)$ token transitions, where $p$ is a prefix of tokens and $v$ is the next token.

Each $(p, v)$ pair is assigned a token-level advantage based on the expected change in reward if $v$ is appended to $p$:
\begin{equation}
A(v \mid p) = \mathbb{E}[R \mid p \circ v] - \mathbb{E}[R \mid p]
\end{equation}

We approximate these expectations using empirical averages over the sampled completions. Let $\mathcal{G}$ be the set of all sampled completions for a given prompt. We define
\begin{align}
\mathbb{E}[R \mid p] &= \frac{1}{|\mathcal{G}_p|} \sum_{y \in \mathcal{G}_p} R(y), \\
\mathbb{E}[R \mid p \circ v] &= \frac{1}{|\mathcal{G}_{p \circ v}|} \sum_{y \in \mathcal{G}_{p \circ v}} R(y),
\end{align}
where $\mathcal{G}_p$ is the subset of completions in which $p$ appears as a prefix, and $\mathcal{G}_{p \circ v}$ is the subset in which the prefix is extended by token $v$. 

The objective used to update the policy is:
\begin{align}
\mathcal{J}_{\text{TSPMO}}(\theta) &= \frac{1}{|\mathcal{U}|} \sum_{(p,v) \in \mathcal{U}} \pi_\theta(v \mid p) \cdot A(v \mid p) \nonumber \\
&\quad - \lambda \sum_{W \in \mathcal{W}_{\text{LoRA}}} \|W\|_2^2
\end{align}
$\mathcal{U}$ is the set of unique $(\text{prefix}, \text{token})$ pairs extracted from all completions, and $\lambda$ is a regularization coefficient applied to the LoRA parameters $\mathcal{W}_{\text{LoRA}}$.

Consistent with our implementation of S-GRPO, we update the policy once per problem, and forego PPO-style objective clipping.

\paragraph*{Replay-Based Resampling}

As most generations will diverge from one another near the start of the generation, we introduce a configurable replay mechanic that serves to build prefix tries in later sections of completions. This allows the model to learn strategies past the beginning of generation.

Let $\mathcal{R}$ denote the set of all previously generated completions, each associated with a scalar reward $R$. We define a threshold value $r \in \mathbb{R}$ and classify completions into:
\begin{itemize}
    \item \textbf{Successful completions:} $\mathcal{R}_\text{success} = \{ y \in \mathcal{R} \mid R(y) \geq r \}$
    \item \textbf{Failed completions:} $\mathcal{R}_\text{fail} = \{ y \in \mathcal{R} \mid R(y) < r \}$
\end{itemize}

At each update step, we may configure replay to sample from either subset. Given a user-defined replay budget of $C_{\text{success}}$ and $C_{\text{failure}}$ completions, we select $C_{\text{success}}$, $C_{\text{failure}}$ completions from $\mathcal{R}_\text{success}$ and  $\mathcal{R}_\text{fail}$ respectively. For each selected completion, we randomly sample a token position $t$ such that
\[
t \in \left[ T_{\text{prompt}}, \; T_{\text{total}} \right],
\]
where $T_{\text{prompt}}$ is the index of the last token in the input question, and $T_{\text{total}}$ is the final token index of the full completion. This ensures that replay focuses on the model-generated portion of the sequence.

For each selected completion and its sampled replay position $t$, we restart generation from the partial sequence corresponding to the prefix $P = (y_1, y_2, \dots, y_t)$, where $y_t$ is the token at the chosen replay point. From this prefix, we again generate $|\mathcal{G}|$ new completions using the current policy $\pi_\theta$. These completions are then evaluated with the reward function, a new independent set of $(\text{prefix}, \text{token})$ advantage pairs are collected, and finally the policy model is updated again using the T-SPMO objective function.

\subsection{Full GRPO Baseline}

To isolate the impacts of sparse‐token sampling, we implement the
original GRPO objective (Eq.\,\ref{eq:grpo}). We again set the number of iterations per sample to 1. The full GRPO objective is thus equivalent to the S-GRPO objective (Eq. \ref{eq:sgrpo}), except all tokens are included in $\mathcal{T}_i$. We adapt GRPO to low-memory environments using gradient accumulation: We generate $\mathcal{G}$ without gradient enabled, calculate rewards, then for each completion in $\mathcal{G}$ we run a forward and backward pass. 

\section{Experiments and Results}
\subsection{Experiment Setup}

We evaluate our RL methods on two reasoning benchmarks: the SVAMP dataset and a custom arithmetic task consisting of multiplication of two 3 digit numbers. Each dataset is handled as a separate training run, with distinct hyperparameter configurations tuned for the nature of the task.

\paragraph*{Model and Fine-Tuning Method} All experiments use the Qwen2-1.5B model fine-tuned with Low-Rank Adaptation (LoRA), with only adapter parameters updated and base weights frozen. We insert LoRA modules into the query and value projections of the final attention layers: Last 1/3 attention layers, rank = 16 for all SVAMP experiments, last 1/4 attention layers, rank = 8 for all multiplication experiments. Based on early experiments, we used a higher LoRA rank for SVAMP, reflecting its greater semantic variability relative to multiplication. Fine-tuning is performed on a single partitioned A100 40GB GPU using AdamW optimizer with a learning rate of $1 \times 10^{-4}$ and regularization coefficients $\beta = \lambda = 0.01$. Training is conducted in float32 precision.

\paragraph*{Training Configuration}
For both benchmarks, responses are limited to a maximum of 300 tokens, with temperature set to $0.3$. Reward is based on exact match accuracy, where the final integer extracted from the model's output is treated as the prediction. 

We use an effective batch size of 1 unless otherwise noted. Gradient accumulation is disabled by default.

S-GRPO and the baseline GRPO samples $|\mathcal{G}| = 8$ completions per prompt, while T-SPMO uses $|\mathcal{G}| = 50$ completions. 

Other hyperparameter settings were selected from our ablation experiments and listed in the Ablations section.

\paragraph*{Training Duration}
Training steps for each method–dataset pair were selected based on early-stopping criteria (plateau detection) and manual selection of the checkpoint with peak performance for multiplication. Full details are provided in the Ablations section. Training steps for the baseline GRPO were selected to match those of S-GRPO.

\subsection{Results}
We report accuracy on the SVAMP testset and a generated multiplication testset.

\begin{table}[H]
\caption{\textit{SVAMP benchmark test-set (n=300) results showing accuracy for each method.}}
        \label{tab:svamp-results}
\begin{center}
\begin{small}
\begin{sc}
\begin{tabular}{lcccr}
		\toprule
		\textbf{Method}    & \textbf{Accuracy (\%)} \\
		\midrule
		Base Qwen2-1.5B    & 45.0 \\
            GRPO               & 46.7 \\
		S-GRPO             & 70.3 \\
		T-SPMO             & 71.6 \\
		\bottomrule
\end{tabular}
\end{sc}
\end{small}
\end{center}
\vskip -0.1in
\end{table}

Table~{\ref{tab:svamp-results}} shows both S-GRPO and T-SPMO substantially improved performance on SVAMP over the base model. We also find that GRPO did not meaningfully enhance the model's performance compared to the base model.

\begin{table}[H]
\caption{\textit{3 digit x 3 digit multiplication test-set (n=3000) results showing accuracy for each method.}}
        \label{tab:mult-results}
\vskip 0.15in
\begin{center}
\begin{small}
\begin{sc}
\begin{tabular}{lcccr}
		\toprule
		\textbf{Method}    & \textbf{Accuracy (\%)}   \\
		\midrule
		Base Qwen2-1.5B    & 3.9\\
            GRPO               &  4.4 \\
		S-GRPO             & 22.9\\
		T-SPMO             & 70.0\\
		\bottomrule
\end{tabular}
\end{sc}
\end{small}
\end{center}
\vskip -0.1in

\end{table}

As demonstrated in Table~{\ref{tab:mult-results}}, the base Qwen2-1.5B model struggled significantly on 3 digit x 3 digit multiplication. Our baseline GRPO also did not improve performance. We see noticeable improvement after training with S-GRPO, but an early plateau in training resulted in low albeit improved test set accuracy. T-SPMO on the other hand excelled on this benchmark, significantly outperforming the baseline and S-GRPO. 

\begin{figure}[t!]
\centering
\includegraphics[width=1\columnwidth]{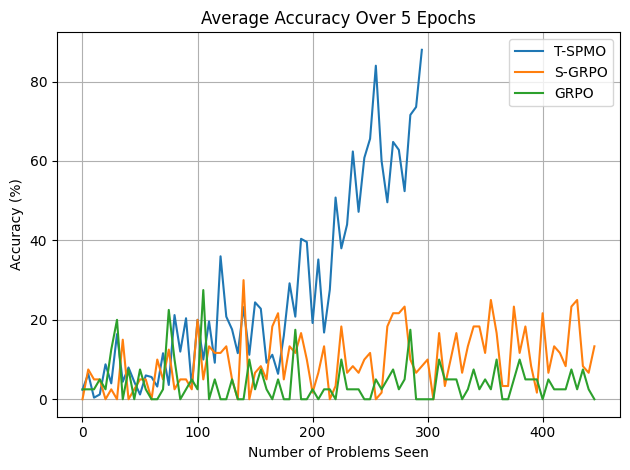}
\caption{Training plots of T-SPMO, S-GRPO, GRPO on multiplication of 3 digit x 3 digit integers}
\label{fig:training-plot}
\end{figure}

We include training curves for the multiplication task in Figure~{\ref{fig:training-plot}}, where the performance divergence between T-SPMO and S-GRPO is very pronounced. 

\begin{table}[H]
\caption{\textit{Average number of tokens per example contributing to the loss during the first 100 epochs of RL training for GRPO, S-GRPO, and T-SPMO. Includes replay tokens for T-SPMO.}}
\label{tab:num-tokens}
\vskip 0.15in
\begin{center}
\begin{small}
\begin{sc}
\begin{tabular}{lcccr}
		\toprule
		\textbf{Method} & \textbf{SVAMP} & \textbf{Mult} \\
		\midrule
		GRPO & 1204.5 & 2166.8 \\
		S-GRPO   & 557.8 & 674.6 \\
		T-SPMO    & 21.0 & 54.1 \\
		\bottomrule
\end{tabular}
\end{sc}
\end{small}
\end{center}
\vskip -0.1in

\end{table}
Table~{\ref{tab:num-tokens}} reports the average number of loss-contributing tokens per example for each method, measured over the first 100 epochs of training. As the model converges, its outputs become more stable and deterministic, leading to fewer loss tokens—especially for methods like T-SPMO. We therefore focus on early training as a practical upper bound on memory usage. Both S-GRPO and T-SPMO substantially reduce the number of optimized tokens, using less than 50\% and 5\%, respectively, of what GRPO consumes.

\subsection{Ablations} We conducted ablations to measure the sensitivity of S-GRPO and T-SPMO to key hyper-parameters. For each algorithm–dataset pair, a pilot sweep with handcrafted settings determined an early-stopping point.

SVAMP: Reward plateauing yielded 400 steps for T-SPMO and 325 for S-GRPO.

3-digit multiplication (more out-of-distribution): we chose the checkpoint with peak validation accuracy after plateauing—350 steps for T-SPMO and 450 for S-GRPO. We also extended T-SPMO training beyond the plateau to verify that additional optimization did not close the performance gap.

All subsequent ablations reused the step counts established for each algorithm–dataset pair to ensure a consistent compute budget and isolate the effect of each hyperparameter.

For S-GRPO, we varied the cutoff $\alpha$ and the maximum number of contributing tokens $k$. The probability parameter $P$ was not ablated independently, as it is inherently tied to $\alpha$ and $k$ --- higher $P$ values are only meaningful when combined higher k and lower $\alpha$. We also do not ablate the group size $|\mathcal{G}|$, as memory usage (or wall-clock time when using gradient accumulation) scales directly with $|\mathcal{G}|$, making larger values infeasible and out of scope to test. Furthermore, $|\mathcal{G}|$ is not unique to S-GRPO and is a shared hyperparameter with GRPO, making it a lower priority for ablation in our study.

\begin{table}[H]
\caption{\textit{S-GRPO ablations on $\alpha$ and $k$ values. Test set accuracies on SVAMP (n=300) and multiplication (n=3000).}}
\label{tab:sgrpo-ablations}
\vskip 0.15in
\begin{center}
\begin{small}
\begin{sc}
\begin{tabular}{lcccr}
		\toprule
		\textbf{($\alpha$, $k$)} & \textbf{SVAMP Acc (\%)} & \textbf{Mult Acc (\%)} \\
		\midrule
		(0, 100)   & 66.0 & 13.0 \\
		(50, 100)  & 70.0 & 22.9 \\
		(100, 100) & 70.3 & 12.7 \\
		(25, 50)   & 68.7 & 12.6 \\
		(0, 25)    & 69.0 & 20.9 \\
		\bottomrule
\end{tabular}
\end{sc}
\end{small}
\end{center}
\vskip -0.1in

\end{table}

Table~{\ref{tab:sgrpo-ablations}} shows that across SVAMP, S-GRPO was relatively robust to changes in $\alpha$ and $k$, with accuracies remaining between 66\% and 70.3\%. 

On the multiplication benchmark, the influence of hyper-parameters was more pronounced, but without a clear pattern. Surprisingly, we find that ($\alpha=50$, $k=100$) and ($\alpha=0$, $k=25$) achieved stronger performance compared to intermediate values. These results suggest that two distinct strategies may be effective for S-GRPO in challenging problem spaces: (1) handcrafting token selection to align with task-specific structure, or (2) conservatively subsampling a small number of tokens across the completion to reduce noise. In contrast, indiscriminately including a large number of tokens without regard to response structure appears to introduce excessive variance and impair optimization.

We also ablated the effective batch size for S-GRPO by accumulating gradients across multiple prompts before each optimization step (with total problems seen held constant).

\begin{table}[H]
\caption{\textit{S-GRPO ablations on effective batch size. Test set accuracies on SVAMP and multiplication benchmarks.}}
\label{tab:sgrpo-batch-ablations}
\vskip 0.15in
\begin{center}
\begin{small}
\begin{sc}
\begin{tabular}{lcccr}
		\toprule
		\textbf{Batch Size} & \textbf{SVAMP Acc (\%)} & \textbf{Mult Acc (\%)} \\
		\midrule
		1  & 70.3 & 22.9 \\
		8  & 62.7 & 15.0 \\
		\bottomrule
\end{tabular}
\end{sc}
\end{small}
\end{center}
\vskip -0.1in
\end{table}

Table~{\ref{tab:sgrpo-batch-ablations}} shows that increasing the effective batch size to 8 led to a significant degradation in performance across both tasks, likely because updates to the policy occur less frequently.

Finally, for T-SPMO, we ablated the group size $|\mathcal{G}|$ and whether completions were replayed based on success or failure.

\begin{table}[H]
\caption{\textit{T-SPMO ablations on group size and replay strategy. Test set accuracies on SVAMP and multiplication benchmarks.}}
\label{tab:tspmo-ablations}
\vskip 0.15in
\begin{center}
\begin{small}
\begin{sc}
\begin{tabular}{lcccr}
		\toprule
		\textbf{($|\mathcal{G}|$, $C_{\text{Suc}}$, $C_{\text{Fail}}$)} & \textbf{SVAMP Acc (\%)} & \textbf{Mult Acc (\%)} \\
		\midrule
		(50, 1, 1) & 71.0 & 62.0 \\
		(50, 1, 0) & 70.0 & 70.0 \\
		(50, 0, 1) & 68.7 & 24.5 \\
		(50, 0, 0) & 71.6 & 19.5 \\
		(25, 1, 1) & 70.0 & 47.8 \\
		(8, 1, 1)  & 69.3 & 11.1 \\
		\bottomrule
\end{tabular}
\end{sc}
\end{small}
\end{center}
\vskip -0.1in
\end{table}

We see in Table~{\ref{tab:tspmo-ablations}} that T-SPMO showed strong performance on SVAMP across all configurations. The multiplication benchmark proved to be more sensitive to hyper-parameters.  Decreasing $|\mathcal{G}|$ consistently reducing performance on multiplication, likely due to insufficient exploration of alternative token transitions during optimization. Running T-SPMO without replay from a successful completion also led to significant performance degradation on multiplication, likely because T-SPMO does not optimize the model for later tokens in completions. Interestingly, for multiplication, including failed completions for replay degraded performance, indicating that forcing the model to recover from early errors may inject too much variance into training.

Overall, these ablations suggest that both S-GRPO and T-SPMO are relatively robust to moderate hyperparameter changes, but benefit from careful tuning on more out-of-distribution tasks.

\subsection{Discussion}

We hypothesize that T-SPMO outperforms S-GRPO on the multi-digit multiplication benchmark due to differences in how each method handles token-level credit assignment and partial correctness. While both encourage chain-of-thought reasoning, S-GRPO applies group-level updates across sampled tokens, making it sensitive to isolated errors: a single mistake can disproportionately penalize an otherwise accurate sequence. 

T-SPMO addresses this by computing advantages for specific (prefix, token) pairs, allowing correct decisions to be reinforced even when later errors occur. 

Extending these insights, our GRPO baseline did not meaningfully improve model performance over the base model on either benchmark. This suggests that GRPO may share the sensitivity issues of S-GRPO in an even more pronounced form, particularly when only a small subset of parameters is updated. We caution against interpreting this as evidence that GRPO is an inferior algorithm broadly; rather, it may overfit under LoRA fine-tuning or require more extensive hyperparameter exploration to stabilize.

Overall, we recommend T-SPMO for tasks where token-level correctness is critical, such as arithmetic reasoning, and S-GRPO for settings where broader output structures matter more than local accuracy. Practitioners should weigh the tradeoffs between selective token-level optimization and group-level coherence depending on the demands of the task.

\section{Conclusions}
We presented two reinforcement learning methods—Stochastic Group Relative Policy Optimization (S-GRPO) and Token-Specific Prefix Matching Optimization (T-SPMO)—designed specifically for LoRA-based fine-tuning of large language models under tight memory and compute constraints. Both methods are critic-free and intentionally reduce the number of output tokens contributing to the gradient, enabling efficient and stable training. By aligning the algorithmic structure with the limitations of low-rank adaptation, our approaches demonstrate that substantial reasoning gains are possible without full-sequence loss computation or heavyweight model updates. Across two benchmarks, SVAMP and multi-digit multiplication, our approaches significantly outperform zero-shot baselines, with T-SPMO achieving particularly strong results on arithmetic tasks.

In addition, our GRPO baseline, which uses full-trajectory loss attribution, did not meaningfully improve model performance over the base model, suggesting that GRPO may exhibit similar sensitivity issues as S-GRPO, but in a more pronounced form when used with parameter-efficient fine-tuning. This highlights the importance of careful reward attribution under constrained settings, and the potential need for additional stabilization or tuning when adapting full-model RL algorithms to LoRA setups.

Our work targets a complementary problem setting: enabling reinforcement learning for reasoning tasks where full fine-tuning is not feasible. We believe this lightweight and efficient approach fills a critical gap for researchers and practitioners working with limited hardware. Future work will explore extending these techniques to more semantically diverse tasks and investigating scaling strategies that preserve the efficiency and fine-grained credit assignment benefits of token-level optimization.

\section*{Impact Statement}

This paper presents work whose goal is to advance the field of 
Machine Learning. There are many potential societal consequences 
of our work, none which we feel must be specifically highlighted here.


\bibliography{example_paper}
\bibliographystyle{icml2025}

\newpage
\appendix
\onecolumn
\section{Prompt Templates}

\subsection{Multiplication Task}
The following template was used for multiplication problems:

\begin{quote}
\texttt{System: You are a helpful AI assistant.}
\\
\texttt{User: What's [Number A] multiplied by [Number B]? Don't use a calculator.}
\\
\texttt{Assistant: Let's break this problem down step by step.}
\\
\end{quote}

Here, \texttt{[Number A]} and \texttt{[Number B]} are placeholders that were replaced with randomly sampled 3-digit integers (between 101 and 999).

\subsection{SVAMP Task}
The following prompt template was used for SVAMP problems:

\begin{quote}
\texttt{Problem: [SVAMP Question]}
\\
\texttt{Answer:}
\end{quote}

In this template, \texttt{[SVAMP Question]} denotes a natural language math word problem drawn from the SVAMP dataset. The model was tasked with generating only the final numerical answer following the "Answer:" cue, without explaining intermediate steps.

\section{Answer Parsing Methodology}

To evaluate model responses, we employed a simple parsing strategy designed to robustly extract the final numerical answer:

\begin{itemize}
    \item \textbf{Comma Removal:} All commas were stripped from the model output to handle large numbers formatted with thousands separators (e.g., ``1,234'' becomes ``1234'').
    \item \textbf{Integer Extraction:} After comma removal, we used a regular expression to locate all integers in the response. Specifically, we selected the \textbf{last integer} found as the predicted answer.
\end{itemize}

This method ensures that extraneous text or intermediate calculations produced by the model do not interfere with accuracy evaluation, yet still mostly isolates correctness over formatting to optimize. Only the final numerical answer is considered for grading.

\end{document}